\documentclass{article}
\usepackage[utf8]{inputenc}

\usepackage{times}
\usepackage{latexsym}
\usepackage{amsmath}
\usepackage{amssymb}
\usepackage{graphicx}
\usepackage{array}
\usepackage{booktabs}
\usepackage{multirow}
\usepackage{rotating}
\usepackage{float}
\usepackage{multicol}
\usepackage[bookmarks=true]{hyperref}
\usepackage[table]{xcolor}
\usepackage{subcaption}
\usepackage{caption}
\usepackage[preprint]{corl_2026}

\usepackage{titlesec}
\usepackage{enumitem}
\usepackage{caption}

\titlespacing*{\paragraph}
{0pt}      
{0.2em}    
{0.5em}    

\setlist[itemize]{
    topsep=1pt,
    itemsep=1pt,
    parsep=0pt,
    partopsep=0pt,
    leftmargin=*
}

\title{LCLA: \textbf{L}anguage-\textbf{C}onditioned \textbf{L}atent \textbf{A}lignment for Vision-Language Navigation}

\author{
  Nitesh Subedi\\
  Iowa State University\\
  Ames, Iowa 50014, USA\\
  \texttt{nitesh@iastate.edu}\\
  \And
  Adam Haroon\\
  Iowa State University\\
  Ames, Iowa 50014, USA\\
  \texttt{aharoon@iastate.edu}\\
  \And
  Samuel Tetteh\\
  Iowa State University\\
  Ames, Iowa 50014, USA\\
  \texttt{samtett@iastate.edu}\\
  \AND
  Prajwal Koirala\\
  Cornell University\\
  Ithaca, New York 14850, USA\\
  \texttt{pk596@cornell.edu}\\
  \And
  Cody Fleming\\
  Iowa State University\\
  Ames, Iowa 50014, USA\\
  \texttt{flemingc@iastate.edu}\\
  \And
  Soumik Sarkar\\
  Iowa State University\\
  Ames, Iowa 50014, USA\\
  \texttt{soumiks@iastate.edu}
}

\begin{document}
\maketitle

\vspace{-8mm}
\begin{figure}[H]
    \centering
    \includegraphics[width=0.95\linewidth]{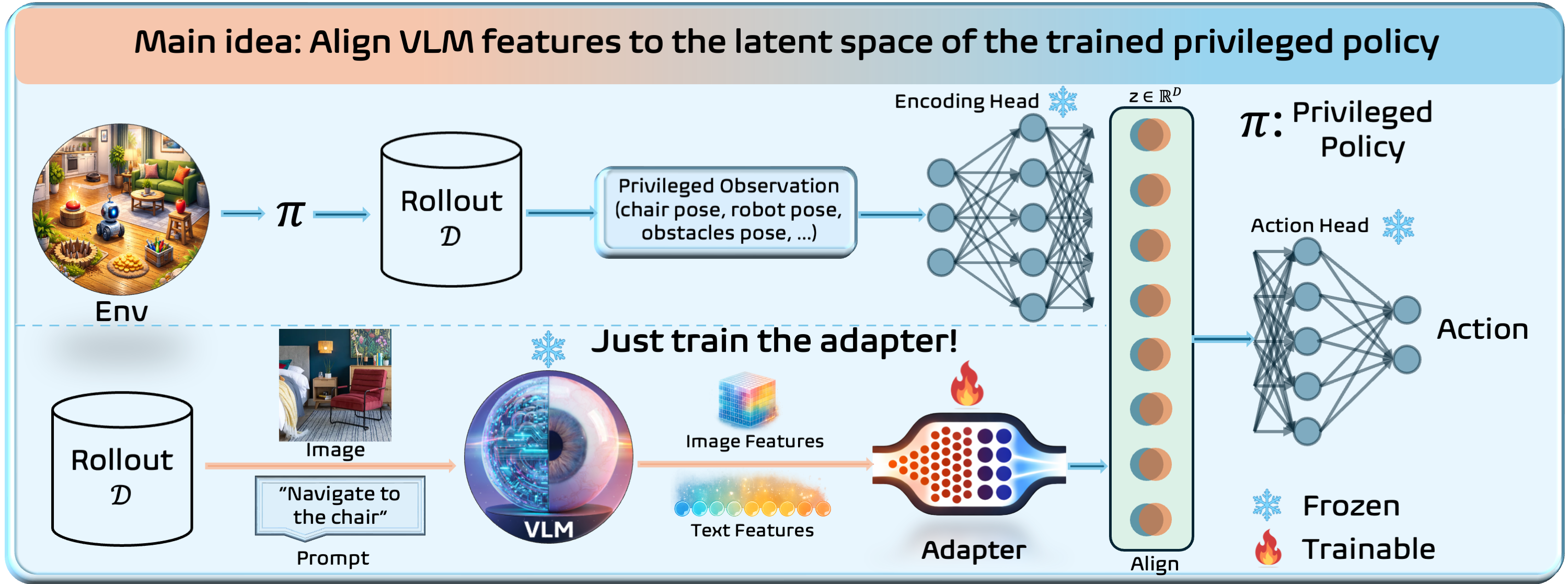}
    \caption{Overview of LCLA. A privileged expert policy trained from geometric state observations induces a task-relevant latent representation $\mathbf{z} \in \mathbb{R}^D$. A lightweight adapter aligns frozen vision-language embeddings to this latent space while keeping the expert policy frozen. During deployment, visual and language inputs are mapped into the expert latent space and passed through the frozen action head for control.}
    \label{fig:abstract}
\end{figure}

\begin{abstract}
We propose LCLA (Language-Conditioned Latent Alignment), a framework for vision-language navigation that decouples perception from control by aligning visual-language observations to the latent representation of a frozen privileged expert policy. A lightweight adapter maps frozen vision-language embeddings into the expert latent space using supervised latent alignment rather than end-to-end policy optimization, enabling expert navigation behavior to be reused across perception backbones and environmental variations. We evaluate LCLA across in-distribution simulation, zero-shot out-of-distribution environments, robustness perturbations, and direct real-world deployment on a TurtleBot platform, where it improves OOD success rate from $49.3\%$ to $80.5\%$ and reduces collision rates from $45.0\%$ to $15.5\%$ relative to direct behavior cloning. Without any real-world fine-tuning or calibration, LCLA attains $70.7\%$ real-world success, compared to $60.9\%$ for a navigation-pretrained baseline and $40.2\%$ for direct behavior cloning, demonstrating robust and efficient sim2real transfer while remaining lightweight at inference time.
\end{abstract}


\section{Introduction}
A central challenge in embodied AI is learning representations that support reliable decision making from high-dimensional sensory inputs~\citep{levine2016endtoendtrainingdeepvisuomotor,Lesort_2018}. Navigation is a canonical instance: control depends on low-dimensional geometric quantities such as goal direction and obstacle proximity, but agents in realistic settings must infer these from raw RGB observations and natural-language instructions. Vision-Language Navigation (VLN) compounds this further~\citep{anderson2018vision,chen2024cvln,huang2022visual,li2025scalable,zhang2024gsa,zhou2024etpnav,zheng2025efficient}: viewpoint changes, lighting variation, and linguistic ambiguity make end-to-end approaches from pixels to actions~\citep{mirowski2016learning,zhu2017target,savva2019habitat} prone to shortcut learning and poor generalization~\citep{agarwal2021deep,geirhos2020shortcut}.

Full navigation requires long-horizon planning, semantic memory, and hierarchical task decomposition, capabilities typically handled by a high-level planner that sequences subgoals while a low-level controller executes them. This paper addresses the low-level component: given a visible target described by a natural-language instruction, navigate toward it reliably. Accordingly, we study language-conditioned object-goal navigation rather than classical long-horizon VLN, isolating the perception-to-control grounding problem from exploration, memory, and subgoal sequencing. Without a robust grounding layer, higher-level planners have nothing stable to hand off to.

We introduce \textbf{Language-Conditioned Latent Alignment (LCLA)}, a framework for this grounding step that decouples perception from control. A privileged expert policy is trained on compact geometric state and then frozen. A vision-language adapter maps image and language instructions into the expert's latent space via supervised alignment, reusing the frozen action head at deployment. This avoids end-to-end policy optimization while leveraging frozen vision-language models~\citep{radford2021learning,zhai2023siglip} for rich perceptual features~\cite{10492662, rls3}. Unlike prior teacher--student methods~\citep{Chen2020LearningByCheating,lu2025contrastiverepresentationlearningrobust,mousa2025tarteacheralignedrepresentationscontrastive} that distill behavior at the action level, LCLA aligns perception to a fixed task-centric latent, enabling the expert controller to be reused across modalities and environments without retraining control dynamics.

We instantiate LCLA as \emph{LCLAA} (LCLA Adapter) for indoor object navigation. Despite training in a single environment, the agent generalizes zero-shot to unseen scenes and objects, and transfers directly to a real TurtleBot with no fine-tuning. We ablate along two axes, latent alignment vs.\ direct action prediction (LCBC), and language-conditioned vs.\ pooled-embedding alignment (PELA), and analyze sensitivity to the latent interface choice. Our contributions are:

\begin{itemize}[noitemsep, topsep=0pt, leftmargin=*, partopsep=0pt]
    \item \textbf{Latent alignment as a grounding primitive:} We reformulate goal-directed vision-language navigation as supervised alignment to a frozen privileged expert's latent space, decoupling perception from control and enabling a single expert controller to be reused across vision-language backbones and environments without retraining.
    \item \textbf{Zero-shot sim2real transfer:} We deploy the resulting policy directly on a TurtleBot with no real-world fine-tuning or calibration, improving success rate and trajectory efficiency over both direct behavior cloning and a navigation-pretrained baseline.
    \item \textbf{Controlled ablations isolating what matters:} Targeted comparisons against action-level imitation (LCBC) and pooled-embedding alignment (PELA), together with sweeps over latent dimensionality and vision-language backbone, identify latent-level supervision and token-level language conditioning as the two components driving robustness under distribution shift.
\end{itemize}

\section{Related Work}

Visual navigation has been studied as learning control directly from high-dimensional sensory observations. Early end-to-end visuomotor approaches map pixels to actions using reinforcement learning~\citep{mirowski2016learning,zhu2017target,savva2019habitat}, but often struggle to generalize under appearance and layout variations due to shortcut learning~\citep{geirhos2020shortcut}. Cognitive mapping methods improve long-horizon reasoning by learning latent spatial representations~\citep{gupta2017cognitive,chaplot2020neural}, though they remain susceptible to perceptual aliasing.

Privileged information frameworks improve training stability by granting experts access to full state information~\citep{pinto2017asymmetric,rajeswaran2016epopt}. Learning by Cheating~\citep{Chen2020LearningByCheating} distills privileged expertise into visual policies through imitation, with related ideas applied to crowd navigation~\citep{monaci2022dipcan} and agile locomotion~\citep{chane2024soloparkour}. Teacher--student methods~\citep{zhangcat,messikommer2025studentinformedteachertraining} and contrastive sim2real approaches~\citep{lu2025contrastiverepresentationlearningrobust,mousa2025tarteacheralignedrepresentationscontrastive} similarly leverage expert supervision, but primarily distill actions or value functions, or durectly clone the actions fom the vlm representations~\cite{prelcla}. LCLA instead aligns observations directly to the expert latent, enabling reuse of frozen control policies without action-level imitation. LCLA is also related to privileged-latent adaptation methods such as RMA~\citep{rma}, which infer dynamics-related context from proprioceptive history for a pretrained controller. In contrast, LCLA predicts an instruction-dependent control latent from RGB and language, such that the same scene may induce different latents for different language-specified goals.

Representation learning aims to improve robustness and sample efficiency in reinforcement learning. Self-supervised objectives~\citep{sermanet2018tcn,laskin2020curl} and data augmentation~\citep{yarats2021drq} improve generalization from pixels, while model-based methods learn compact latent dynamics for planning~\citep{ha2018world,hafner2019dreamer,vjepa2}. Bisimulation-based approaches explicitly target task-relevant abstractions~\citep{zhang2021bisimulation}. However, these representations are typically learned jointly with the policy and remain tightly coupled to the observation modality and training environment. Zhang et al.~\cite{zhang2021roach} distill a privileged RL coach into a monocular driving policy using latent and policy supervision. In contrast, LCLA uses pretrained vision-language representations and language conditioning to map observations into the latent space of a frozen teacher policy, reusing its original control head without retraining.

Vision-language navigation conditions navigation on natural-language instructions~\citep{anderson2018vision,fried2018speaker,krantz2020beyond, 10.3389/frobt.2022.930486}. Large vision-language models such as CLIP and SigLIP-2~\citep{radford2021learning,tschannen2025siglip2} enable zero-shot and language-conditioned navigation~\citep{dorbala2022clip,majumdar2022zson,Shah2022LMNav}, with subsequent work integrating VLM features into spatial maps~\citep{huang2022visual} or improving adaptation in continuous environments~\citep{zhou2024etpnav,zheng2025efficient,zhang2024gsa,chen2024cvln,li2025scalable}. Most approaches treat VLM features as generic perceptual embeddings and optimize downstream policies end-to-end. LCLA instead aligns vision-language observations to a predefined expert latent, enabling modular reuse of frozen control policies across backbones and environments.

\section{Methodology}
\label{sec:method}


Our approach decouples perception from control by separating the agent into:
(i) a privileged expert policy trained in a compact latent control space, and
(ii) a visual-language adapter that maps RGB observations and language instructions into this latent space. The privileged expert policy is denoted:
\[
z_t = \pi^{e}_{\text{priv}}(s_t), \qquad
a_t = \pi^{a}_{\text{priv}}(z_t),
\]
where $s_t$ is a privileged state observation, $z_t$ is an intermediate latent representation, and $a_t$ is the output action. The adapter predicts a corresponding latent directly from visual observations, $\hat{z}_t = f_\theta(I_t, L)$, where $I_t$ is an RGB image and $L$ is a language instruction. At deployment the agent has no access to privileged state: actions are produced by chaining the frozen modules,
\[
I_t, L \xrightarrow[]{f_\theta} \hat{z}_t
\xrightarrow[]{\pi^{a}_{\text{priv}}} a_t.
\]

\subsection{Privileged Expert Policy}

We first train a privileged PPO policy using low-dimensional geometric state variables such as goal direction, obstacle distances, and relative pose information. These states are invariant to appearance, texture, and lighting variations, enabling the policy to learn task-centric navigation behavior independent of visual observations.

After training, the expert encoder $\pi^{e}_{\text{priv}}$ and action head $\pi^{a}_{\text{priv}}$ are frozen. We use the intermediate latent representation $z_t$ as the target interface for latent alignment.

\subsection{Visual-Language Latent Alignment}

We roll out the frozen privileged expert while simultaneously rendering RGB observations, recording tuples $(I_t, L, z_t)$ at each timestep. A frozen vision-language backbone extracts patch-level visual embeddings $V_t$ and text embedding $T$, and a lightweight adapter predicts $\hat{z}_t = f_\theta(V_t, T)$ via supervised alignment to $z_t$. No reinforcement learning is performed at this stage; both the expert policy and VLM backbone remain frozen throughout. The adapter architecture is described in Section~\ref{subsec:lclaa}.

Our implementation uses a lightweight language-conditioned attention adapter (Fig.~\ref{fig:lclaa}) that selectively attends to task-relevant visual features before projecting them into the expert latent space.


\subsection{LCLAA: Language-Conditioned Latent Alignment Adapter}
\label{subsec:lclaa}

We validate the proposed latent alignment framework using a specific adapter instantiation termed \textbf{LCLAA} (Language-Conditioned Latent Alignment Adapter). While LCLA is architecture-agnostic, LCLAA provides a concrete realization that maps frozen vision-language representations into the latent space of a frozen privileged expert policy.

\begin{figure*}[h]
\centering 
\includegraphics[width=\linewidth]{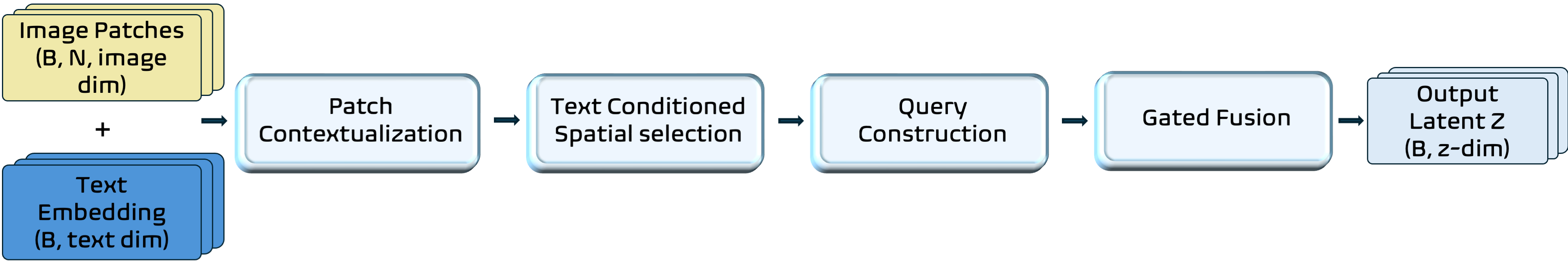} 
\caption{\textbf{Architecture of the Language Conditioned Latent Alignment Adapter (LCLAA)}. Image patches and text embeddings are processed through language-conditioned attention blocks to select task-relevant visual information and predict the aligned latent representation $\hat{z}_t$.}
\label{fig:lclaa} 
\end{figure*}

Given RGB observation $I_t$ and language instruction $L$, a frozen vision-language backbone encodes patch-level visual embeddings $V_t = \phi_{\mathrm{img}}(I_t)$ and text embedding $T = \phi_{\mathrm{txt}}(L)$. The adapter predicts an expert-compatible latent representation $\hat z_t = f_\theta(V_t, T)$, which is passed through the frozen privileged policy action head for control.

\paragraph{Expert latent interface:}
The privileged PPO policy uses a multilayer architecture
$\mathrm{Priv\_obs} \rightarrow 512 \rightarrow 256 \rightarrow 128 \rightarrow \mathrm{action}$. We treat the intermediate activations of this policy as candidate latent interfaces. For each candidate layer, we train a separate vision--language adapter with latent dimension $d_z \in \{512,256,128\}$, while keeping the corresponding downstream PPO sub-network frozen during deployment. We further perform an ablation study over these latent interfaces to evaluate alignment quality and downstream navigation performance, as discussed in Section~\ref{sec:results}.

\paragraph{Adapter architecture:}
LCLAA uses language-conditioned attention to selectively extract task-relevant visual information before latent projection (Fig.~\ref{fig:lclaa}). Patch embeddings are first contextualized through self-attention, after which instruction-conditioned attention weights are used to aggregate semantically relevant visual regions:
\[
\bar x_t = \sum_{i=1}^{N} \alpha_{t,i} x_{t,i},
\]
where $\alpha_{t,i}$ denotes the language-conditioned attention weight for patch $x_{t,i}$.

The aggregated visual representation is subsequently refined through stacked cross-attention blocks that fuse visual and textual context. Finally, a lightweight MLP projects the fused representation into the expert latent space:
\[
\hat z_t = \mathrm{MLP}_\theta(\mathrm{Fuse}(Q_t, T)).
\]

This design enables the adapter to preserve spatially relevant task semantics while aligning frozen vision-language representations with the privileged control latent space.

\paragraph{Training objective:}
Let $z_t$ denote the privileged expert latent. The adapter is trained using a combined latent and action alignment objective:
\[
\mathcal{L}
=
\lambda_1 \mathcal{L}_{\mathrm{contrast}}
+
(1-\lambda_1)\|\hat z_t - z_t\|_2^2
+
\lambda_2
\|
\pi_{\mathrm{priv}}^{a}(\hat z_t)
-
\pi_{\mathrm{priv}}^{a}(z_t)
\|_2^2.
\]

The terms provide complementary supervision: latent MSE constrains absolute representation error, while symmetric InfoNCE over $\ell_2$-normalized latents ($\tau=0.07$) encourages relational alignment between student and expert representations. Action consistency addresses a distinct failure mode: latents that are close in representation space can still induce substantially different actions if their discrepancy lies along a control-sensitive direction. It therefore penalizes latent errors according to their downstream effect under the frozen controller.

Only adapter parameters are optimized; both the vision-language backbone and privileged policy remain frozen throughout training. In our case, we used $\lambda_1=0.8$ and $\lambda_2=1$.

\paragraph{Training details:} We collect $52$k expert rollout steps by executing the frozen PPO policy in the training environment. To improve robustness under limited environment diversity, we augment visual observations offline using aggressive appearance transformations. Specifically, for each original observation, we generate four additional augmented views using random combinations of cropping, color jitter, contrast and sharpness changes, blurring, perspective distortion, noise injection, and random erasing. This results in an effective training set of approximately $260$k visual--language samples (one original and four augmented views per rollout step). The adapter is then trained in a fully supervised manner for $10$ epochs to align visual--language embeddings to the corresponding expert latent representations. Unless stated otherwise, all results use SigLIP2-Base as the backbone. 

\begin{figure}
    \centering
    \includegraphics[width=0.9\linewidth]{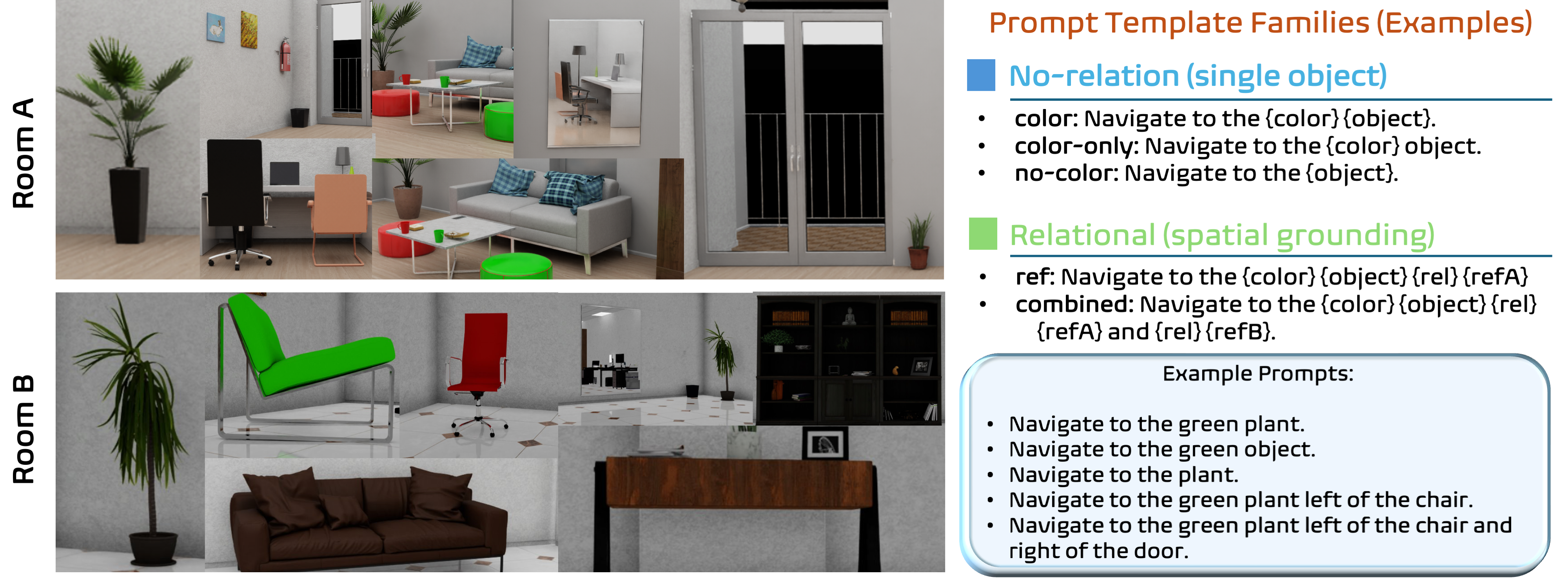}
    \caption{The left panels show example Room~A and Room~B environments with varied objects and layouts, while the right panel summarizes the language prompt templates used for adapter training. Linguistic and spatial variation encourages compositional grounding and OOD generalization.}
    \label{fig:rooms_fig}
\end{figure}

\subsection{Experimental Setup}

\paragraph{Task:}
We evaluate all methods on an object-goal navigation task. At the beginning of each episode, the agent is initialized at a random pose in an indoor environment and receives a language instruction $L$ describing a target object category (e.g., ``Navigate to the chair''). An episode is considered successful if the agent reaches within a predefined distance threshold of any valid target instance while approximately facing the object within a fixed time horizon.

To encourage semantic grounding beyond category labels, each target object is associated with multiple language prompts containing diverse contextual descriptions. These prompts include appearance-based attributes (e.g., color) as well as relational descriptions referencing nearby objects and scene context. During training, prompts are randomly sampled for each episode. Example environments and prompt variations are shown in Figure~\ref{fig:rooms_fig}.

\paragraph{Simulation Environments:}
Experiments are conducted in photorealistic indoor environments built using Isaac Sim/Isaac Lab~\citep{mittal2025isaaclab}. We construct two environment splits to evaluate both in-distribution (ID) navigation performance and zero-shot out-of-distribution (OOD) generalization.

Room A serves as the training environment and contains 18 target objects spanning common household categories such as chairs, cabinets, and tables. Object instances, layouts, and scene appearance are fully observed during training. Room B is held out entirely during training and used exclusively for evaluation. It contains 12 target objects, including both visually novel instances of seen categories and entirely unseen object categories. In addition to semantic novelty, Room B differs substantially in spatial arrangement, textures, lighting conditions, and overall scene geometry. As a result, successful navigation in Room B requires semantic grounding and spatial generalization rather than memorization of layouts or object appearance.

\begin{figure}[h]
    \centering
    \includegraphics[width=\linewidth]{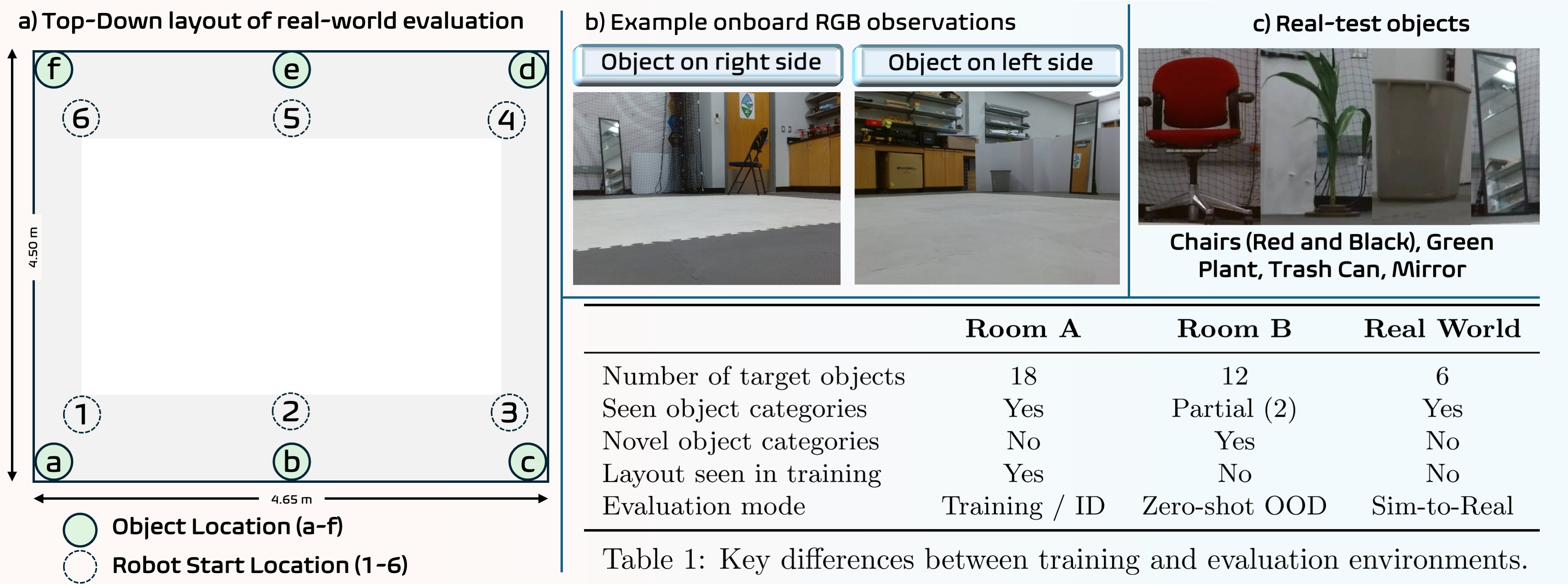}
    \caption{Real-world evaluation setup. (a) Top-down layout of the evaluation arena showing robot start positions (1--6) and target object locations (a--f). (b) Example onboard RGB observations for left- and right-side object placements. (c) Example real-world target objects used during evaluation.}
    \label{fig:lab_experiment}
\end{figure}

\paragraph{Real-World Evaluation:}
To evaluate sim2real transfer, we deploy all policies directly on a TurtleBot platform with no real-world fine-tuning, calibration, or adaptation; the robot operates solely from onboard RGB and language instructions. The arena (Fig.~\ref{fig:lab_experiment}) defines six start positions and six target locations spanning two background regimes (positions 1--3 share context; 4--6 introduce distinct backgrounds and clutter). We evaluate six office object instances seen during training (chairs, plants, and trash bins of varying color). Each scene contains the goal object and a distractor of a different category or appearance, so success requires grounding the instruction rather than approaching the most salient object. We additionally evaluate semantic robustness using same-category objects with varying appearances and semantics. Each method is tested under left- and right-biased object placements, resulting in 82 real-world trials per method. Full details are in Fig.~\ref{fig:lab_experiment} and Appendix~\ref{app:realworld}.

\paragraph{Baselines:} We compare LCLA against representative end-to-end methods and targeted ablations:
\begin{itemize}
    \item \textbf{Privileged Policy}: A PPO policy trained directly from privileged state observations to actions \citep{schulman2017ppo}. This policy is used to collect expert trajectories and serves as an upper bound for in-distribution performance (Room A).

    \item \textbf{Language-Conditioned Behavior Cloning (LCBC)}: 
    This baseline removes latent alignment and directly predicts expert actions from visual–language observations. It uses the same backbone and encoder architecture as LCLA, but replaces the latent adapter and frozen action head with a trainable action predictor (an MLP mapping $256 \rightarrow 128 \rightarrow \text{action}$), matching the structure of the expert action head. This controlled ablation isolates the effect of latent alignment by changing only the prediction target from latent representations to actions.

    \item \textbf{Pooled-Embedding Latent Alignment (PELA)}: PELA ablates explicit language conditioning by aligning expert latents from \emph{pooled} vision-language embeddings rather than token-level representations. Global image and text features from the frozen VLM are concatenated and mapped to the expert latent via an MLP ($\text{embeddings}\!\rightarrow\!512\!\rightarrow\!512\!\rightarrow\!\hat z$). The perception backbone, latent target, and objective are unchanged, isolating the contribution of explicit language conditioning.
    
    \item \textbf{LeLaN}: A language-conditioned navigation policy trained on large-scale augmented egocentric data using foundation models for supervision \citep{hirose2024lelan}. Although not fine-tuned on our environment, LeLaN benefits from navigation-specific vision-language supervision, whereas our approach uses off-the-shelf vision-language encoders, making the comparison conservative in favor of LeLaN.
\end{itemize}

\paragraph{Controlled ablation protocol.}
To isolate the contributions of perception and control representation, we evaluate a $2\times2$ design crossing pooled versus language-conditioned vision--language features with direct-action versus privileged-latent prediction. PEBC, the pooled direct-action variant, predicts expert actions from pooled SigLIP2 image and text features; together with LCBC, PELA, and LCLA, it completes the four configurations. Room~A is used for controlled component isolation, while Room~B provides a zero-shot generalization check. We additionally evaluate a capacity-matched PELA variant and ablations of the LCLA training objective; full results are provided in Appendix~\ref{app:additional_ablations}.

\paragraph{Metrics:} We report \textbf{Success Rate} (SR) and \textbf{SPL} (Success weighted by Path Length). SR measures whether the goal is reached; SPL measures efficiency given the shortest path. All results are averaged over random seeds and episode initializations (30 episodes per goal for simulation).

\section{Results}
\label{sec:results}

\setcounter{table}{1}
\begin{table*}[h]
\centering
\small
\setlength{\tabcolsep}{3.0pt}
\begin{tabular}{lccc|ccc|cc|cc}
\toprule
\multirow{2}{*}{\textbf{Method}} 
& \multicolumn{3}{c|}{\textbf{Room A (ID)}} 
& \multicolumn{3}{c|}{\textbf{Room B (OOD)}} 
& \multicolumn{2}{c|}{\textbf{Real World}}
& \multicolumn{2}{c}{\textbf{Efficiency}} \\
\cmidrule(lr){2-4}
\cmidrule(lr){5-7}
\cmidrule(lr){8-9}
\cmidrule(lr){10-11}

& \textbf{SR} & \textbf{SPL} & \textbf{Coll.} 
& \textbf{SR} & \textbf{SPL} & \textbf{Coll.}
& \textbf{SR} & \textbf{SPL}
& \textbf{Params (Train./Total)} & \textbf{Time(ms)} \\
\midrule

Privileged PPO 
& 97.7 & 0.977 & 2.3 
& -- & -- & -- 
& -- & --
& 0.49M / 0.49M & 5.0 \\

\midrule

LeLaN 
& 61.1 & 0.606 & 33.7 
& 54.8 & 0.548 & 41.1 
& 60.9 & --
& 30.5M / 181.9M & 135.7 \\

LCBC 
& 60.1 & 0.591 & 39.7 
& 49.3 & 0.491 & 45.0 
& 40.2 & 0.297
& 13.7M / 99.7M & 75.4 \\

PELA
& 62.5 & 0.616 & 37.5
& 57.6 & 0.576 & 40.9
& -- & --
& \textbf{1.2M / 87.2M} & \textbf{69.1} \\

LCLA 
& \textbf{90.4} & \textbf{0.903} & \textbf{9.6} 
& \textbf{80.5} & \textbf{0.804} & \textbf{15.5} 
& \textbf{70.7} & \textbf{0.522}
& 13.0M / 99.0M & 74.2 \\

\bottomrule
\end{tabular}
\vspace{1mm}
\caption{Performance across in-distribution (Room A), zero-shot OOD (Room B), and real-world navigation settings. SR denotes Success Rate, SPL denotes Success weighted by Path Length, and Coll.\ denotes collision rate (\%). Real-world evaluation is performed without fine-tuning over 82 trials per method. Proxy SPL is estimated from executed trajectory length when trajectory information is available. PELA is evaluated in simulation only.}
\label{tab:full_results}
\end{table*}

We evaluate LCLA across in-distribution simulation (Room~A), zero-shot out-of-distribution evaluation (Room~B), and direct real-world deployment on a TurtleBot platform. Table~\ref{tab:full_results} reports Success Rate (SR), Success weighted by Path Length (SPL), collision rate, parameter counts, and inference time for all methods. Table 3 presents ablations on latent dimensionality and backbone robustness under illumination and viewpoint perturbations.

\paragraph{In-distribution performance (Room A).}
The privileged PPO expert achieves $97.7\%$ SR, confirming the task is solvable given geometric state and establishing a practical upper bound. Among deployable methods, LCLA substantially outperforms all baselines, reaching $90.4\%$ SR within $7\%$ of this bound despite operating from raw RGB and language. LCBC, PELA, and LeLaN all cluster near $60$--$62\%$ SR with $3$--$4\times$ higher collision rates, far below LCLA, even though LeLaN uses a backbone $\sim\!14\times$ larger with navigation-specific pretraining. LCLA additionally runs at nearly half LeLaN's inference latency (Table~\ref{tab:full_results}).

\paragraph{Zero-shot out-of-distribution performance (Room B).}
Room~B is held out entirely during training and introduces unseen object categories, novel layouts, lighting, and geometry. LCLA retains $80.5\%$ SR, compared with $49.3\%$ for LCBC, $57.6\%$ for PELA, and $54.8\%$ for LeLaN. To further disentangle perception and control representation, we also evaluate a pooled-embedding behavior-cloning baseline (PEBC), which directly predicts actions from pooled SigLIP2 image and text features; PEBC achieves $49.0\%$ SR / $0.479$ SPL in Room~A and $19.4\%$ / $0.194$ in Room~B. Together, PEBC, LCBC, PELA, and LCLA form a controlled $2\times2$ comparison between pooled versus language-conditioned perception and direct-action versus privileged-latent prediction, showing that both components contribute to generalization. LCLA also reduces the Room~B collision rate from $45.0\%$ for LCBC to $15.5\%$.
\begin{figure}[h]
    \centering
    \captionsetup{labelformat=empty}
    \includegraphics[width=\linewidth]{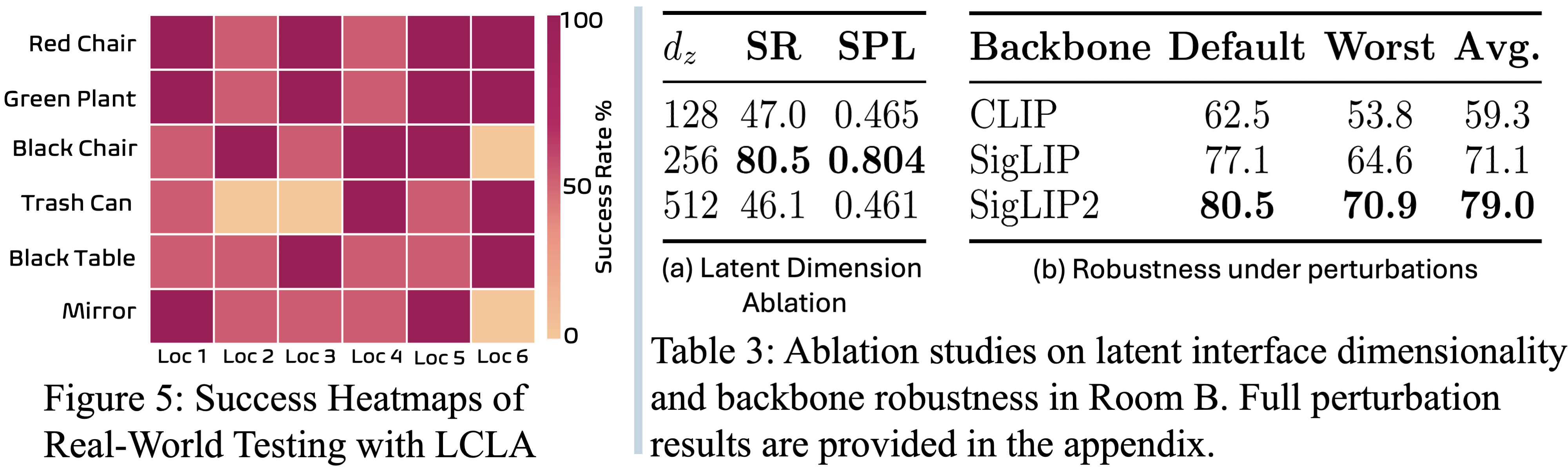}
    \caption{}
    \label{fig:heatmaps}
    \vspace{-5mm}
\end{figure}

\begin{figure}[t]
    \centering
    \includegraphics[width=0.5\linewidth]{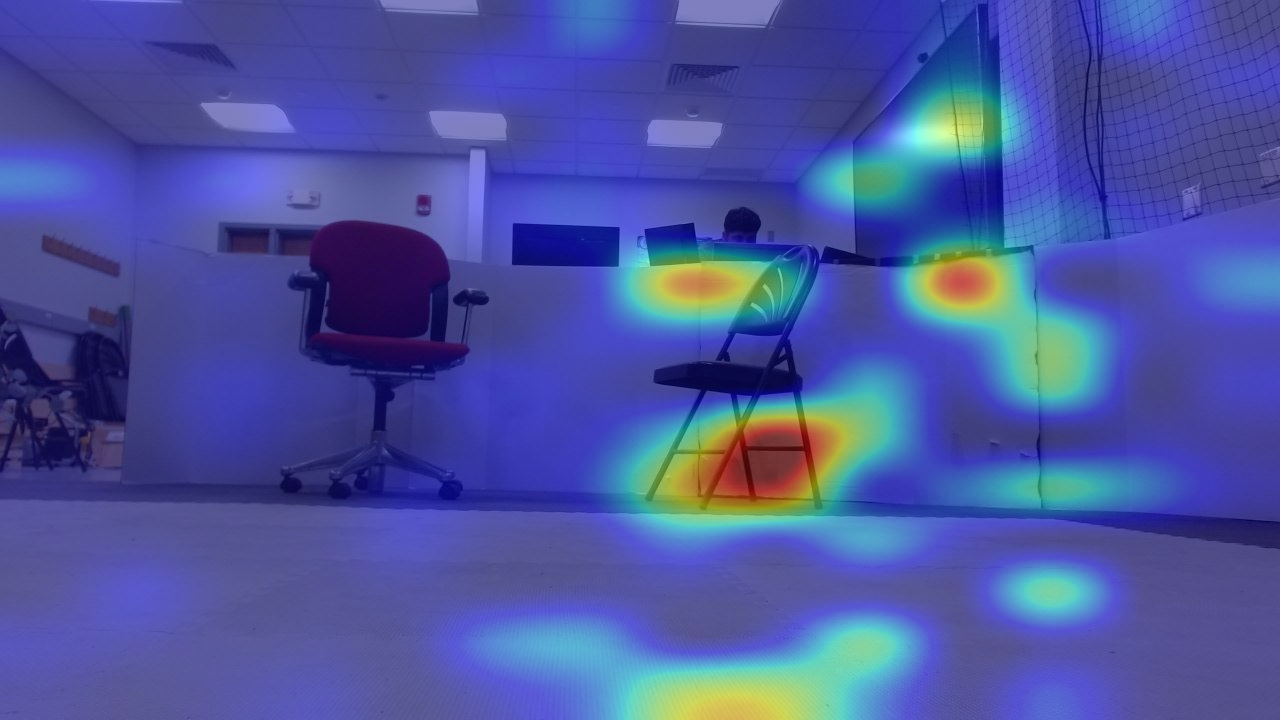}
    \caption{Attention map for "Navigate to the black chair."}
    \label{fig:attention_map}
\end{figure}

\paragraph{Zero-shot real-world deployment.}
All policies are deployed directly on a TurtleBot with no real-world fine-tuning, policy calibration, or domain adaptation; the robot operates solely from onboard RGB and the language instruction. LCLA achieves the highest real-world success rate ($70.7\%$, proxy SPL $0.522$), outperforming both LeLaN ($60.9\%$) and LCBC ($40.2\%$) despite using off-the-shelf vision-language encoders rather than navigation-specific pretraining. The gap to LCBC is particularly stark: LCBC suffers catastrophic failures at several arena locations, repeatedly colliding or stalling, which indicates that the action-level supervision used by direct behavior cloning does not transfer the spatial structure required for reliable real-world execution. LCLA, in contrast, maintains consistent performance across start positions and background regimes, suggesting that alignment to the expert latent preserves control-relevant structure robustly enough to absorb sensor noise, lighting variation, and actuation uncertainty without sim2real adaptation. Detailed real-world results, including performance breakdowns across object categories, arena locations, and start configurations, are provided in the Appendix. An example attention map in real world is shown in Fig.~\ref{fig:attention_map}.

\paragraph{Additional controlled ablations.}
The $2\times2$ results above indicate complementary contributions from language-conditioned perception and latent-level control. Relative to PEBC, language conditioning alone improves Room-A SR by $11.1$ points and the privileged-latent interface by $13.5$ points, while combining both improves SR by $41.4$ points, corresponding to a positive interaction of $16.8$ points. A capacity-matched pooled latent adapter improves PELA from $62.5\%$ to $71.8\%$ SR but remains below LCLA, showing that adapter size explains only part of the gap. Objective ablations yield $49.0\%$ SR with latent MSE alone, $57.0\%$ with MSE+action consistency, $51.0\%$ with MSE+InfoNCE, and $90.4\%$ with the full objective. Finally, linear probes show that $\hat z$ does not explicitly reconstruct the geometric variables encoded by the privileged latent, while the two representations agree on steering direction in $84.3\%$ of held-out states, suggesting preservation of control-relevant rather than explicitly geometric structure. Full results are provided in Appendix~\ref{app:additional_ablations}.

\paragraph{Latent dimensionality and backbone robustness ablation:}
Table 3 reports the effect of latent dimension and backbone on OOD performance. Matching the adapter output to the expert latent dimension ($d_z=256$) is critical: both $d_z=128$ ($47.0\%$ SR) and $d_z=512$ ($46.1\%$) perform far below the default, confirming that stable alignment requires the adapter output to remain consistent with the expert representation manifold. Across backbones, stronger vision-language priors improve alignment robustness: SigLIP2 consistently outperforms SigLIP and CLIP under all perturbations, while degradation remains gradual rather than catastrophic, consistent with the low Lipschitz constant of the frozen action head (Appendix~\ref{app:stability}).

\paragraph{Latent interface selection.} The latent interface is selected empirically through latent-dimension and architecture ablations across models of varying depth and width. Intermediate representations consistently outperform both earlier and deeper alternatives, suggesting a more stable interface between perception and control. In particular, matching the adapter output dimension to the expert latent dimension ($d_z=256$) is critical for stable latent alignment, while both smaller and larger latent spaces lead to degraded success rate and SPL. We further analyze robustness of the frozen latent-action mapping using Lipschitz sensitivity estimates and latent noise-injection experiments, which show gradual degradation under perturbations. Full details are in Appendix~\ref{app:latent} and~\ref{app:stability}.

\section{Discussion}
\label{sec:discussion}

Our results indicate that robust goal-directed navigation can be recovered by aligning perception to a frozen task-centric latent interface rather than learning actions end-to-end. LCBC is unstable under distribution shift, PELA only partially closes the gap, and LCLA improves success, collision avoidance, and efficiency across all settings.

LCLA addresses a specific layer of the navigation problem: given a target object visible in the frame and described by language, navigate to it reliably. A higher-level planner must still decide which goal to pursue, when to switch objectives, and how to sequence subgoals across a scene. Within this scope, the zero-shot real-world results suggest LCLA is stable enough to serve as a low-level primitive: it handles sensor noise, lighting variation, and actuation uncertainty without catastrophic failure, and remains consistent across viewpoints where LCBC breaks down. This stability is encouraged by the latent-alignment design: by matching the expert’s intermediate representation, visual errors are filtered through a frozen action head with low local sensitivity near the expert manifold. The probe results further suggest that this compatibility need not correspond to explicit reconstruction of the teacher's geometric variables; instead, $\hat z$ appears to retain the structure required by the downstream controller. As a result, bounded latent misalignment tends to degrade actions gradually rather than abruptly, while perception updates remain decoupled from the expert controller.

\section{Limitations}
\label{sec:limitations}

\begin{itemize}[noitemsep, topsep=0pt, leftmargin=*, partopsep=0pt]
    \item \textbf{Goal visibility assumption:} LCLA operates as a low-level grounding primitive and assumes the target is visible and reachable from the current pose. Exploration, active search, and multi-room planning are outside its scope. A practical deployment would pair LCLA with a higher-level planner responsible for goal generation and subgoal sequencing.
    \item \textbf{Environment scale:} Experiments are conducted in constrained indoor environments with short-horizon object-goal tasks. Performance in large-scale, highly dynamic, or outdoor settings remains untested, and the current formulation does not address long-horizon planning or semantic memory.
    \item \textbf{Expert policy dependency:} Alignment quality is bounded by the structure of the privileged expert latent. A poorly trained or suboptimal expert produces a latent space that is harder to align to, limiting what the adapter can recover regardless of backbone quality. Consistent with this limitation, under a matched Room-A protocol a weaker expert and its corresponding LCLA student both achieve $79.0\%$ SR / $0.772$ SPL, indicating that student performance tracks the quality of the frozen expert.
\end{itemize}

\section{Conclusion}
\label{sec:conclusion}

We introduced Language-Conditioned Latent Alignment (LCLA), a framework that decouples perception from control by aligning visual--language observations to a frozen expert latent space. Rather than learning actions end-to-end, LCLA recovers task-relevant structure through supervised latent alignment, enabling a frozen expert policy to be reused across perception backbones, environments, and sensing modalities. Across simulation and real-world experiments, LCLA consistently outperforms direct BC and navigation-pretrained baselines, with strong robustness under distribution shift, viewpoint variation, and zero-shot sim2real transfer, all without real-world fine-tuning.

The results position latent alignment as a viable low-level grounding primitive for modular navigation stacks. The most natural extension is integrating LCLA within a hierarchical system where a language model or spatial planner handles goal generation and subgoal sequencing, while LCLA executes individual navigation goals. Beyond this, future work will explore richer language grounding for relational and underspecified instructions, memory-augmented variants for navigation when the target is not initially visible, and evaluation in larger-scale and outdoor environments.

\section{Acknowledgments}

This work is jointly funded by the NSF–USDA COALESCE program (NSF award \#1954556; USDA award \#2021-6702134418), NSF CNS-2313104 and the Iowa Space Grant Consortium under NASA Award No. 80NSSC25M7126.

\clearpage

\bibliography{references}  

\newpage
\appendix

\section{Real-World Evaluation Protocol}
\label{app:realworld}

This appendix details the real-world evaluation arena and trial protocol summarized in Section~\ref{sec:results}.

\paragraph{Arena and object placement.}
The evaluation arena consists of six predefined robot start locations and six possible target object locations (Fig.~\ref{fig:lab_experiment}(a)). For each trial, the target object is placed at one of the six object locations while the robot is initialized at one of the six start positions. To increase viewpoint diversity and reduce positional bias, each configuration is evaluated twice, placing the target object on the left and right side of the robot camera frame (Fig.~\ref{fig:lab_experiment}(b)).

\paragraph{Objects and distractors.}
We evaluate six office object instances commonly observed during training, including multiple chair instances, plants, and trash bins with varying colors and appearances. To increase semantic ambiguity and discourage trivial object memorization, each scene always contains at least two visible objects: a goal object specified by the language instruction and a distractor object of a different category or appearance. Successful navigation therefore requires grounding the instruction to the correct target rather than approaching the most visually salient object.

\paragraph{Background variation.}
The arena contains background variation across locations. Positions 1--3 share similar surrounding backgrounds and scene context, whereas positions 4--6 introduce substantially different visual backgrounds, clutter configurations, and viewing conditions, evaluating robustness to contextual appearance changes in addition to viewpoint shifts.

\section{Additional Real-World Results}
\begin{figure}[H]
    \centering
    \includegraphics[width=0.8\linewidth]{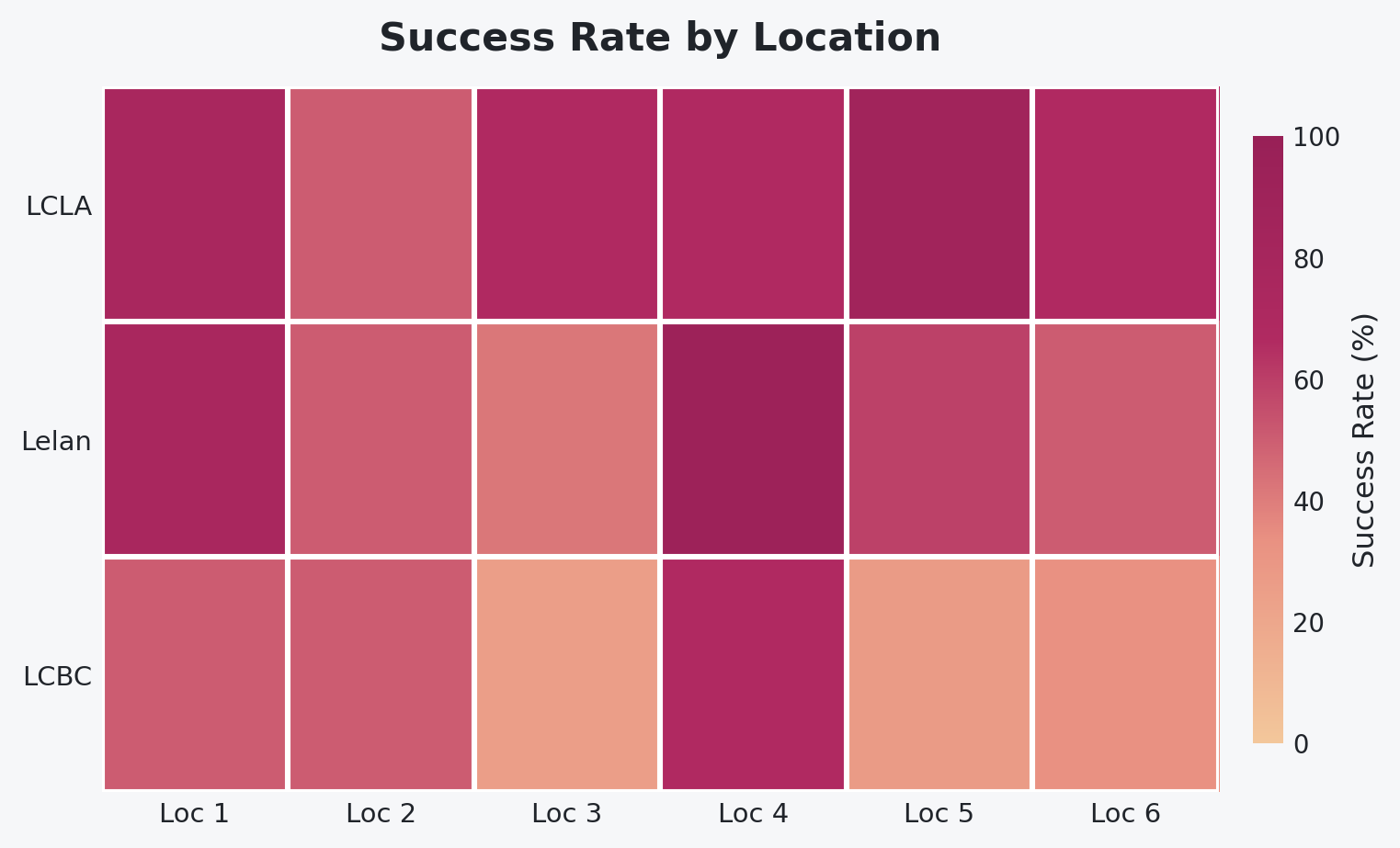}
    \caption{Success rate by object location. Heatmap showing task success rate across evaluated locations for each method. Darker cells indicate higher success rates, highlighting how performance varies with spatial
  placement in the scene.}
    \label{fig:placeholder}
\end{figure}

\begin{figure}[H]
    \centering
    \includegraphics[width=0.8\linewidth]{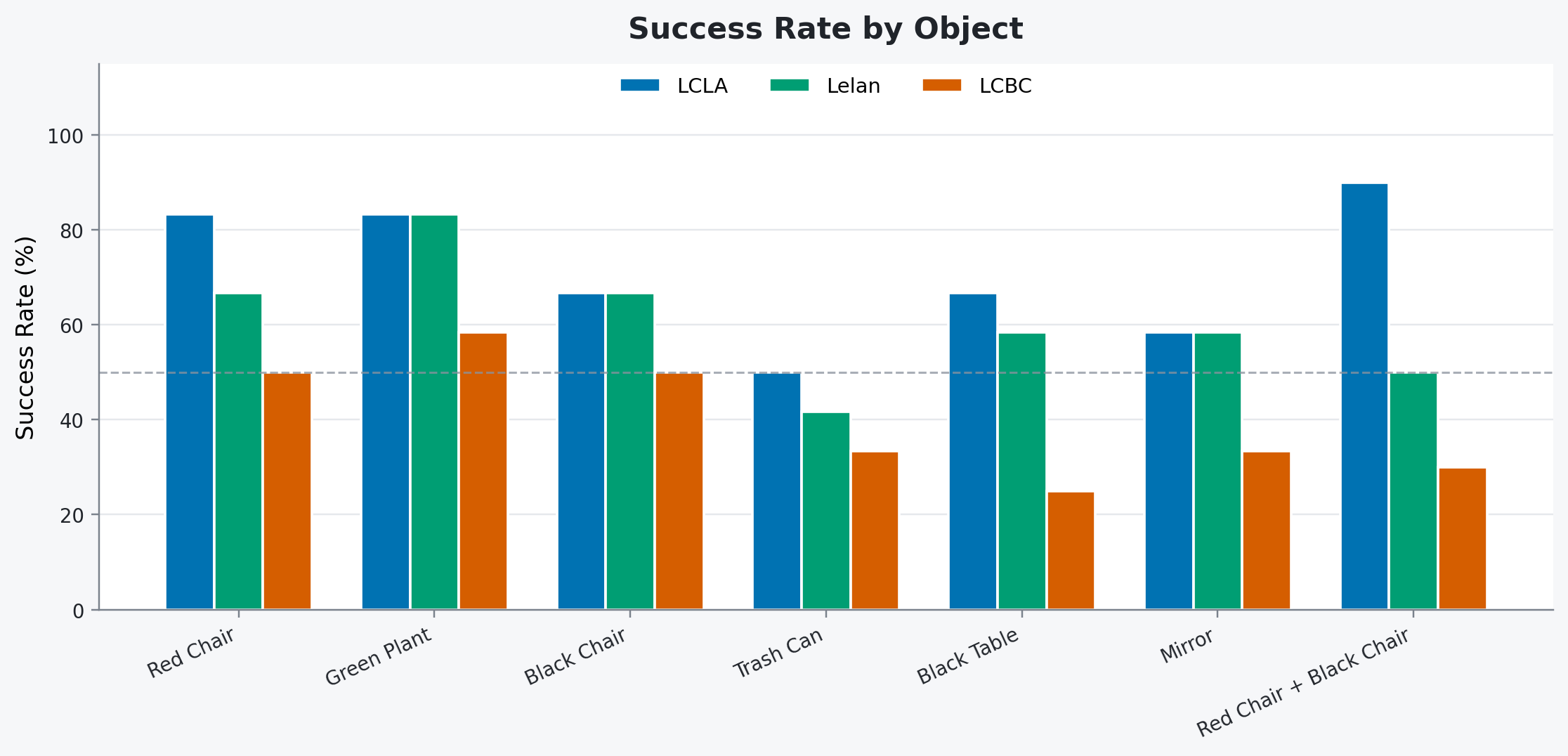}
    \caption{Success rate by target object. Object-wise success rate for each method across all evaluated trials. The plot compares how reliably each method identifies or localizes different objects, including both
  single-object and paired-object targets.}
    \label{fig:placeholder}
\end{figure}

\section{Additional Quantitative Results}

\begin{table}
\centering
\small
\setlength{\tabcolsep}{6pt}
\begin{tabular}{lccc}
\toprule
\textbf{Latent Dimension (Room B)} 
& \textbf{$d_z$}
& \textbf{SR (\%)} 
& \textbf{SPL} \\
\midrule
Reduced latent           & 128 & 47.0 & 0.465 \\
Default latent           & 256 & \textbf{80.5} & \textbf{0.804} \\
Expanded latent          & 512 & 46.1 & 0.461 \\
\bottomrule
\end{tabular}
\caption{Effect of latent dimensionality on out-of-distribution performance. All rows correspond to the same LCLAA agent with different output latent dimensions. Matching the adapter output dimension to the expert latent space ($d_z = 256$) is critical for stable latent alignment; both smaller and larger dimensions lead to severe performance degradation.}
\label{tab:latent_dim_full}
\end{table}

\begin{table}[H]
\centering
\small
\setlength{\tabcolsep}{6pt}
\begin{tabular}{llcc}
\toprule
\textbf{Backbone} & \textbf{Condition (Room B)} 
& \textbf{SR (\%)} 
& \textbf{SPL} \\
\midrule
\multirow{7}{*}{CLIP}
 & Default OOD             & 62.5 & 0.625 \\
 & Very low light          & 53.8 & 0.537 \\
 & Low light               & 63.9 & 0.638 \\
 & High light              & 59.6 & 0.596 \\
 & Very high light         & 61.3 & 0.613 \\
 & Lowered camera ($-0.2$\,m) & 56.4 & 0.564 \\
 & Raised camera ($+0.2$\,m)  & 57.4 & 0.572 \\
\midrule
\multirow{7}{*}{SigLIP}
 & Default OOD             & 77.1 & 0.765 \\
 & Very low light          & 64.6 & 0.643 \\
 & Low light               & 71.3 & 0.707 \\
 & High light              & 71.9 & 0.716 \\
 & Very high light         & 75.0 & 0.746 \\
 & Lowered camera ($-0.2$\,m) & 65.4 & 0.651 \\
 & Raised camera ($+0.2$\,m)  & 72.4 & 0.720 \\
\midrule
\multirow{7}{*}{SigLIP2}
 & Default OOD             & \textbf{80.5} & \textbf{0.804} \\
 & Very low light          & 70.9 & 0.704 \\
 & Low light               & 78.5 & 0.784 \\
 & High light              & \textbf{84.8} & \textbf{0.845} \\
 & Very high light         & 81.4 & 0.814 \\
 & Lowered camera ($-0.2$\,m) & 79.7 & 0.795 \\
 & Raised camera ($+0.2$\,m)  & 82.2 & 0.822 \\
\bottomrule
\end{tabular}
\caption{Backbone-dependent robustness of LCLA to illumination and camera-pose perturbations in Room~B. While all backbones degrade under extreme conditions, stronger vision-language representations (SigLIP, SigLIP2) exhibit substantially improved robustness to viewpoint and lighting shifts. All backbones are base models. The perturbation conditions are illustrated in Figs.~\ref{fig:light_intensity} and~\ref{fig:cam_heights}.}
\label{tab:backbone_full}
\end{table}

\begin{figure*}[h]
    \centering
    \includegraphics[width=\linewidth]{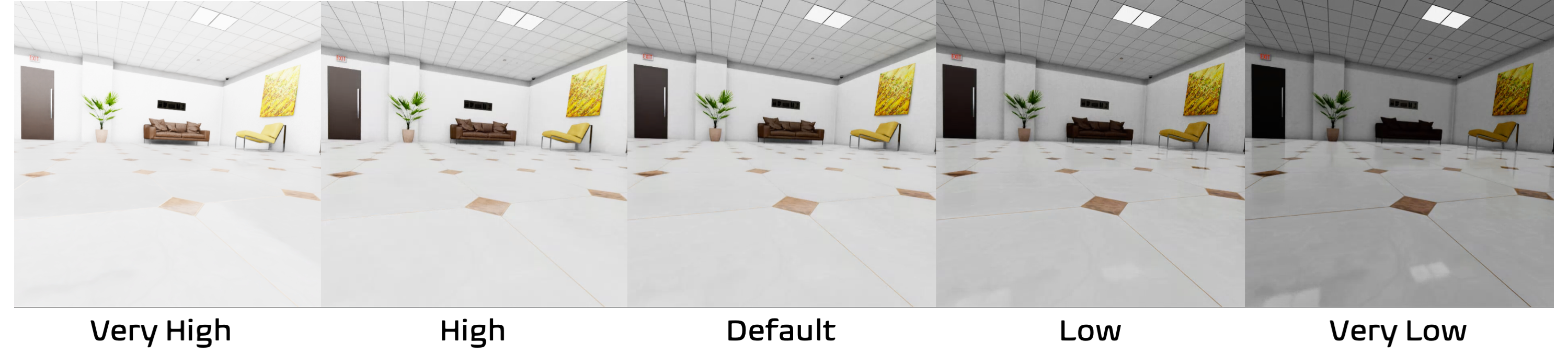}
    \caption{Illumination intensity perturbations in the out-of-distribution Room~B environment. Egocentric RGB observations are shown under very high, high, default, low, and very low lighting conditions. These controlled changes in global illumination significantly alter scene appearance, contrast, and shadowing, and are used to evaluate robustness to lighting variation in Table~\ref{tab:backbone_full}.}
    \label{fig:light_intensity}
\end{figure*}

\begin{figure*}[h]
    \centering
    \includegraphics[width=\linewidth]{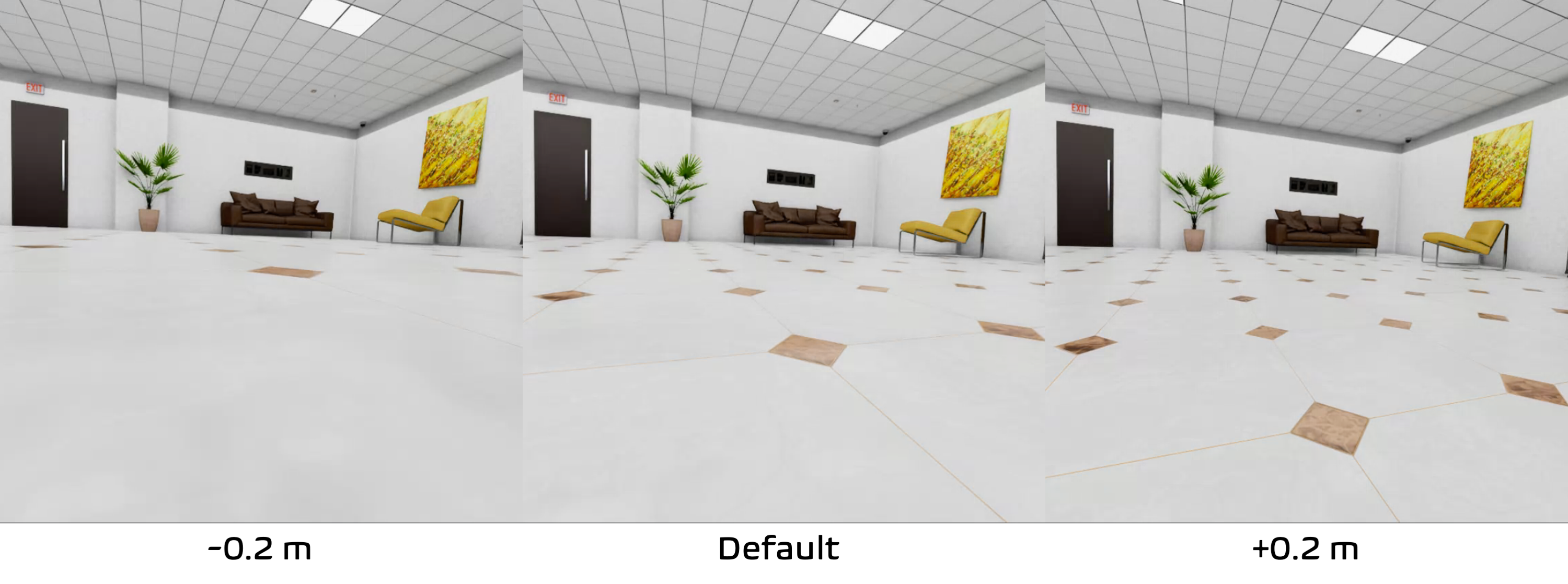}
    \caption{Camera height perturbations in the out-of-distribution Room~B environment. The agent's egocentric RGB observations are shown for a lowered camera ($-0.2$\,m), the default camera height, and a raised camera ($+0.2$\,m). These perturbations induce significant changes in visible floor area, object scale, and perspective, serving as a controlled test of viewpoint robustness in Table~\ref{tab:backbone_full}.}
    \label{fig:cam_heights}
\end{figure*}

\section{Additional Controlled Ablations}
\label{app:additional_ablations}

\subsection{Perception and Control Representation}

To separately evaluate the contributions of language-conditioned perception and the privileged-latent control interface, we complete a $2\times2$ comparison between pooled versus language-conditioned representations and direct-action versus latent prediction. The pooled direct-action baseline (PEBC) uses pooled SigLIP2 image and text features and directly predicts expert actions.

\begin{table}[H]
\centering
\small
\setlength{\tabcolsep}{5pt}
\begin{tabular}{llcccc}
\toprule
\textbf{Perception} & \textbf{Prediction}
& \multicolumn{2}{c}{\textbf{Room A}}
& \multicolumn{2}{c}{\textbf{Room B}} \\
\cmidrule(lr){3-4}
\cmidrule(lr){5-6}
& & SR (\%) & SPL & SR (\%) & SPL \\
\midrule
Pooled & Action
& 49.0 & 0.479 & 19.4 & 0.194 \\

Language-conditioned & Action
& 60.1 & 0.591 & 49.3 & 0.491 \\

Pooled & Latent
& 62.5 & 0.616 & 57.6 & 0.576 \\

Language-conditioned & Latent
& \textbf{90.4} & \textbf{0.903}
& \textbf{80.5} & \textbf{0.804} \\
\bottomrule
\end{tabular}
\caption{Controlled $2\times2$ comparison between perception representation and control interface.}
\label{tab:factorial_ablation}
\end{table}

Relative to pooled direct-action prediction, language-conditioned perception improves Room-A SR by $11.1$ points, while the privileged-latent/frozen-head interface improves SR by $13.5$ points. Combining both components improves SR by $41.4$ points, corresponding to a positive interaction of $16.8$ points.

\subsection{Capacity-Matched PELA}

Because the original PELA adapter contains fewer trainable parameters than LCLA, we additionally increase only the capacity of the pooled-feature adapter while keeping the pooled SigLIP2 input, privileged-latent target, and frozen action head unchanged. Capacity-matched PELA improves from $62.5\%$ SR / $0.616$ SPL to $71.8\%$ / $0.716$ in Room~A. Thus, adapter capacity explains part of the original gap, although the capacity-matched model remains $18.6$ SR points below LCLA.

\subsection{Training-Objective Ablation}

\begin{table}[H]
\centering
\small
\begin{tabular}{lcc}
\toprule
\textbf{Objective} & \textbf{SR (\%)} & \textbf{SPL} \\
\midrule
Latent MSE & 49.0 & 0.486 \\
Latent MSE + Action MSE & 57.0 & 0.567 \\
Latent MSE + InfoNCE & 51.0 & 0.505 \\
Full objective & \textbf{90.4} & \textbf{0.903} \\
\bottomrule
\end{tabular}
\caption{Ablation of the LCLA training objective in Room~A.}
\label{tab:loss_ablation}
\end{table}

Latent regression anchors the predicted representation to the privileged latent, InfoNCE provides relational alignment, and action consistency constrains errors according to their downstream control effect. Action consistency provides the larger individual improvement over latent MSE ($+8.0$ versus $+2.0$ SR points), while the full objective performs substantially better than either partial objective.

\subsection{Latent Representation Probes}

We train episode-disjoint linear probes on $53{,}556$ randomized rollout states from $215$ episodes to examine what information is linearly accessible from pooled SigLIP2 features, the predicted latent $\hat z$, and the privileged latent $z_{\mathrm{priv}}$.

The privileged latent makes navigation geometry almost perfectly linearly accessible: $R^2=1.000$ for relative goal $x/y$ and goal distance, $R^2=0.989$ for minimum obstacle distance, and the mean goal-bearing error is $0.03^\circ$. In contrast, the corresponding scalar $R^2$ values for $\hat z$ remain near zero, indicating that the predicted latent does not explicitly reconstruct the geometric coordinate structure of the teacher.

To test whether $\hat z$ nevertheless preserves control-relevant information, we pass both $\hat z$ and $z_{\mathrm{priv}}$ through the frozen action head on held-out states. Their actions agree on steering direction $84.3\%$ of the time, with left- and right-wheel correlations of $0.56$ and $0.60$, respectively. These results suggest that $\hat z$ preserves control-relevant structure without explicitly reconstructing the privileged geometric variables.

\subsection{Sensitivity to Privileged-Expert Quality}

We additionally evaluate LCLA using a weaker privileged expert. Under the matched Room-A evaluation protocol, the weaker expert achieves $79.0\%$ SR / $0.772$ SPL, and the corresponding LCLA student also achieves $79.0\%$ / $0.772$. This indicates that the quality of the frozen privileged expert remains a practical limitation, since the student cannot recover control capability absent from the teacher.

\section{Choice of Latent Interface}
\label{app:latent}

The latent interface ($d_z = 256$) was selected by training a separate adapter for each candidate hidden layer of the privileged PPO policy ($d_z \in \{128, 256, 512\}$) and comparing the resulting downstream navigation performance, reported in Table~\ref{tab:latent_dim_full}. Only $d_z = 256$ yields successful navigation; both shallower ($d_z = 128$) and deeper ($d_z = 512$) interfaces collapse to roughly $47\%$ SR despite identical adapter architectures, training data, and frozen action heads. Appendix~\ref{app:stability} provides a complementary sensitivity analysis of the selected interface (Lipschitz bound on the frozen action head and noise-injection behavior). We interpret the sharp $d_z$ dependence as consistent with the role of intermediate layers as an information bottleneck that suppresses nuisance variation while preserving control-relevant structure, although this interpretation is offered as a post-hoc explanation rather than verified by independent diagnostic measurements.




\section{Latent Robustness and Action-Space Stability}
  \label{app:stability}

  \paragraph{Sensitivity of the frozen action head.}
  A potential concern with latent alignment is that imperfect visual latents may move away from the expert latent manifold. We analyze this by estimating the sensitivity of the frozen action head $
  \pi^a_{\mathrm{priv}}$:
  \[
  \|\pi^a_{\mathrm{priv}}(z_1)-\pi^a_{\mathrm{priv}}(z_2)\| \le L\|z_1-z_2\|.
  \]
  For the saved action head, the product of spectral norms gives a conservative global upper bound of $L \le 0.77$ under the $\ell_2$ norm. This indicates that errors in the aligned latent space are not strongly
  amplified by the final action mapping.

  \paragraph{Noise-injection experiments.}
  We further test robustness by injecting isotropic Gaussian noise into expert latents at deployment. Performance degrades gradually rather than catastrophically: success remains $100\%$ at $\sigma=1.0$, drops to
  $88\%$ at $\sigma=2.0$, and remains $75.5\%$ at $\sigma=3.0$. These results suggest that moderate latent perturbations do not immediately destabilize the policy.

  \paragraph{Decoupling perception from control.}
  Latent alignment separates perception from control: the visual adapter learns to match the expert latent representation, while the action head remains fixed. This allows the visual backbone or adapter to be updated
  without retraining the expert controller, provided the resulting latents remain aligned with the expert manifold.

\section{PPO Teacher Training Configuration}
\label{app:ppo_config}

This appendix summarizes the configuration used to train the \emph{privileged PPO teacher} in Isaac Lab for the Jetbot navigation task. We report the environment design choices, observation/action spaces, domain randomization parameters, reward shaping terms, and termination conditions. (Implementation-specific function and file names are omitted.)

\subsection{Environment and Simulation Settings}
\label{app:sim_settings}

\paragraph{Vectorized simulation.}
Training uses $N_{\text{env}}=16$ parallel environments with environment spacing of 50 units and replicated physics enabled. The control decimation is 2, with a simulation step of $\Delta t = 1/60$\,s, yielding an effective control step of $2/60$\,s. Episodes are capped at 15\,s.

\paragraph{Physics and rendering.}
The simulation uses GPU dynamics with a TGS solver and GPU broadphase. Rendering is configured in performance mode with RTX features enabled (global illumination, reflections, shadows, denoisers), using 1 sample per pixel. Ambient light intensity is set to zero. A dome HDR environment light is used (HDRI texture, intensity 1000).

\paragraph{Robot actuation.}
The robot is controlled as a differential-drive system via direct wheel \emph{velocity targets}. The action scale is 40.0 (rad/s). Actuator configuration uses velocity limits of 20.0 (rad/s), zero stiffness, and damping 10.0, emphasizing velocity-control dynamics over position-control behavior.

\subsection{Goal Set and Reset Strategy}
\label{app:goals_reset}

\paragraph{Goal set.}
A discrete set of semantic goals is defined (e.g., door, desk, TV, chairs, plants, lamp, trash can, etc.). Each goal may have one or more candidate $(x,y)$ target coordinates. On reset, a goal is sampled uniformly and a specific variant is selected (closest variant to the start pose when multiple are available).

\paragraph{Room bounds and safe goal adjustment.}
The navigable region is bounded by a fixed axis-aligned box:
\[
(x_{\min},y_{\min},x_{\max},y_{\max}) = (-4.25,\,-3.64,\,6.5,\,3.5).
\]
A ``safe'' goal position is computed by moving the goal slightly toward the room center by a configurable clearance and then clamping to the room bounds with a small margin, preventing targets from landing outside valid space or too close to edges.

\paragraph{Start pose.}
The robot resets to a fixed start position (shared across environments), and the yaw is set to face the current goal with added uniform noise of $\pm 0.5$ radians (approximately $\pm 28.6^\circ$), improving robustness to initial heading perturbations.

\subsection{Observation Space}
\label{app:obs_space}

\paragraph{Privileged PPO observation.}
The PPO teacher uses a privileged state vector of dimension 159, composed of:
\begin{itemize}
    \item \textbf{Base navigation state} (9 dims):
    \[
    [g_x, g_y, \sin(\beta), \cos(\beta), v_x, v_y, \omega, \dot{q}_L, \dot{q}_R],
    \]
    where $(g_x,g_y)$ is the goal position in the robot body frame, $\beta$ is the bearing angle to the goal, $(v_x,v_y)$ is planar velocity in body frame, $\omega$ is yaw rate, and $(\dot{q}_L,\dot{q}_R)$ are left/right wheel velocities.
    \item \textbf{Obstacle features} (150 dims): $K=50$ obstacle clusters, each represented by three body-frame quantities
    \[
    [o_x, o_y, d] \in \mathbb{R}^3,
    \]
    normalized by a maximum obstacle distance $d_{\max}=5.0$\,m. Total obstacle feature dimensionality is $50 \times 3 = 150$.
\end{itemize}

\paragraph{Obstacle clustering.}
Obstacle centers are extracted from a 2D occupancy map generated on the USD stage (cell size 0.05\,m) within fixed bounds. Occupied cells are clustered into $K=50$ centroids via $k$-means, yielding a compact, fixed-size obstacle representation.

\paragraph{Normalization.}
Obstacle coordinates and distances are normalized by $d_{\max}$ before concatenation. Goal coordinates are represented in the robot body frame, and bearing is encoded with $\sin/\cos$ to avoid angle discontinuities.

\paragraph{Optional sensory logging.}
RGB and depth camera outputs (single tiled camera) are captured for evaluation/logging and stored as auxiliary outputs; they are not used as privileged PPO observations in this configuration.

\subsection{Action Space and Control}
\label{app:action_space}

The action space is continuous with dimension 2:
\[
a = [a_L, a_R] \in [-1, 1]^2,
\]
scaled by $s_a = 40.0$ to produce wheel velocity targets in rad/s:
\[
\dot{q}^{\star} = s_a \cdot a.
\]
These targets are applied directly to the wheel joints each control step.

\subsection{Domain Randomization}
\label{app:domain_randomization}

Domain randomization is applied \emph{on reset} via event terms.

\paragraph{Material randomization.}
A set of apartment object prim paths (furniture, electronics, decorations, small objects, structural elements, and door) are randomized by:
\begin{itemize}
    \item sampling a diffuse display color uniformly from a default RGB range $[0.1,0.9]^3$,
    \item unbinding existing materials on mesh children so the display color is visible,
    \item (optionally) randomizing roughness in $[0.2,0.8]$ and metallic in $[0.0,0.3]$.
\end{itemize}
To preserve task semantics, objects corresponding to the \emph{currently selected goal} are excluded from material randomization in each environment.

\paragraph{Lighting randomization (optional).}
Support is included to randomize directional light intensity, angle, and orientation, as well as renderer auto-exposure settings. In the provided configuration, an HDR dome light is active; directional-light randomization can be enabled similarly when used.

\subsection{Reward Function}
\label{app:reward}

The reward is a sum of shaped components designed to provide dense learning signal while enforcing safety:

\paragraph{(1) Progress and potential-based shaping.}
Let $d_t$ be current distance to goal and $d_{t-1}$ the previous distance. Progress reward:
\[
r_{\text{prog}} = (d_{t-1} - d_t), \qquad \lambda_{\text{prog}} = 15.0.
\]
A distance potential term provides a smooth shaping signal:
\[
r_{\text{pot}} = -\lambda_{\text{pot}}\Bigl(1 - e^{-d_t/5}\Bigr)\cdot 5, \qquad \lambda_{\text{pot}} = 2.0.
\]

\paragraph{(2) Heading alignment.}
Let $\hat{h}$ be the unit heading direction and $\hat{g}$ be the unit direction to the goal. The facing reward uses exponential shaping:
\[
r_{\text{face}} = \lambda_{\text{face}}\left(e^{\langle \hat{h},\hat{g}\rangle} - 1\right),
\qquad \lambda_{\text{face}} = 5.0.
\]

\paragraph{(3) Forward motion when aligned.}
Forward motion is rewarded only when the robot is oriented toward the goal ($\langle \hat{h},\hat{g}\rangle > 0$):
\[
r_{\text{fwd}} = \lambda_{\text{fwd}}\; \|\mathbf{v}_{xy}\|\; \max(0, \langle \hat{h},\hat{g}\rangle),
\qquad \lambda_{\text{fwd}} = 0.5.
\]

\paragraph{(4) Obstacle avoidance penalty.}
Obstacle distances are computed in world coordinates from clustered centers. Obstacles near the goal (within 1.5\,m) and obstacles extremely close to the robot (within 0.3\,m) are excluded from penalty computation. Let $d_{\min}$ be the minimum remaining obstacle distance, and $d_{\text{th}} = 0.5$\,m. The penalty is:
\[
r_{\text{obs}} = \lambda_{\text{obs}} \cdot \mathrm{clip}\!\left(\frac{d_{\text{th}} - d_{\min}}{d_{\text{th}}}, 0, 1\right),
\qquad \lambda_{\text{obs}} = -3.0.
\]

\paragraph{(5) Boundary penalty.}
If the robot leaves the room bounds, a fixed penalty is applied:
\[
r_{\text{bnd}} = \lambda_{\text{bnd}} \cdot \mathbb{I}[\text{out of bounds}],
\qquad \lambda_{\text{bnd}} = -10.0.
\]

\paragraph{(6) Success bonus.}
Success is declared when distance and heading are within thresholds:
\[
d_t < d_{\text{succ}} = 1.5\ \text{m}, \qquad |\theta_{\text{err}}| < \theta_{\text{succ}} = \pi/6.
\]
A large terminal bonus is added:
\[
r_{\text{succ}} = \lambda_{\text{succ}}\cdot \mathbb{I}[\text{success}],
\qquad \lambda_{\text{succ}} = 200.0.
\]

\paragraph{Total reward.}
The final reward is:
\[
r_t = \lambda_{\text{prog}} r_{\text{prog}} + r_{\text{pot}} + r_{\text{face}} + r_{\text{fwd}} + r_{\text{obs}} + r_{\text{bnd}} + r_{\text{succ}}.
\]

\subsection{Termination and Truncation}
\label{app:termination}

Episodes terminate upon:
\begin{itemize}
    \item \textbf{Success:} meeting the distance and angle thresholds above.
    \item \textbf{Hard collision:} when the maximum contact force magnitude exceeds a small threshold ($10^{-5}$ as configured).
    \item \textbf{Timeout:} reaching the episode length limit (15\,s).
\end{itemize}
Timeout is treated as truncation only when neither success nor collision has occurred.

\subsection{Sensors and Auxiliary Signals}
\label{app:sensors}

A contact sensor is enabled on the robot to measure net forces used for collision termination. A forward-facing RGB camera is instantiated for evaluation-time logging.

\end{document}